\documentclass[sigconf]{acmart}
\usepackage{xcolor} 
\usepackage{booktabs}
\usepackage{adjustbox}
\usepackage{multirow}
\usepackage{makecell}
\AtBeginDocument{%
  }

\setcopyright{acmlicensed}
\copyrightyear{2018}
\acmYear{2018}
\acmDOI{XXXXXXX.XXXXXXX}
\acmConference[Conference acronym 'XX]{Make sure to enter the correct
  conference title from your rights confirmation email}{June 03--05,
  2018}{Woodstock, NY}
\acmISBN{978-1-4503-XXXX-X/2018/06}

\begin{document}

\title{Think Twice Before You Judge: Mixture of Dual Reasoning Experts for Multimodal Sarcasm Detection}

\author{Soumyadeep Jana}
\email{sjana@iitg.ac.in}
\orcid{0009-0002-1936-4229}
\affiliation{%
 \institution{Indian Institute of Technology Guwahati}
 \city{Guwahati}
 \state{Assam}
 \country{India}
}
 
\author{Abhrajyoti Kundu}
\email{abhrajyoti00@gmail.com}
\orcid{0009-0006-9236-3678}
\affiliation{%
 \institution{Indian Institute of Technology Guwahati}
 \city{Guwahati}
 \state{Assam}
 \country{India}
}
 
\author{Sanasam Ranbir Singh}
\email{ranbir@iitg.ac.in}
\orcid{0000-0003-0484-2144}
\affiliation{%
 \institution{Indian Institute of Technology Guwahati}
 \city{Guwahati}
 \state{Assam}
 \country{India}
}

\renewcommand{\shortauthors}{Soumyadeep Jana, Abhrajyoti Kundu, and Sanasam Ranbir Singh}
\begin{abstract}

Multimodal sarcasm detection has attracted growing interest due to the rise of multimedia posts on social media. Understanding sarcastic image-text posts often requires external contextual knowledge, such as cultural references or commonsense reasoning. However, existing models struggle to capture the deeper rationale behind sarcasm, relying mainly on shallow cues like image captions or object-attribute pairs from images. To address this, we propose \textbf{MiDRE} (\textbf{Mi}xture of \textbf{D}ual \textbf{R}easoning \textbf{E}xperts), which integrates an internal reasoning expert for detecting incongruities within the image-text pair and an external reasoning expert that utilizes structured rationales generated via Chain-of-Thought prompting to a Large Vision-Language Model. An adaptive gating mechanism dynamically weighs the two experts, selecting the most relevant reasoning path. Unlike prior methods that treat external knowledge as static input, MiDRE selectively adapts to when such knowledge is beneficial, mitigating the risks of hallucinated or irrelevant signals from large models. Experiments on two benchmark datasets show that MiDRE achieves superior performance over baselines. Various qualitative analyses highlight the crucial role of external rationales, revealing that even when they are occasionally noisy, they provide valuable cues that guide the model toward a better understanding of sarcasm.

\end{abstract}




\maketitle


\begin{figure}[t]
\centering
\includegraphics[width=0.41\textwidth]{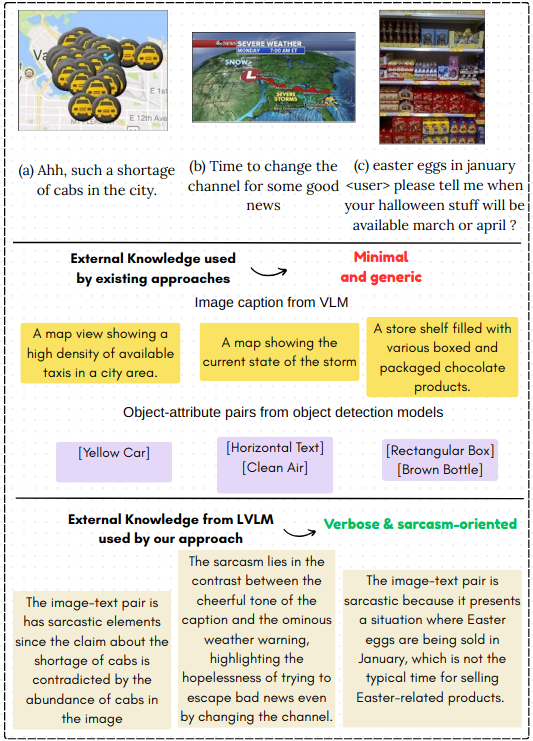}
\caption{Comparison between external knowledge used by existing approaches vs our approach.}
\Description{Diagram comparing external knowledge used by prior works versus our approach.}
\label{fig:intro}
\end{figure}

\section{Introduction}
Sarcasm is a complex figure of speech where the intended meaning of an utterance often differs from the literal meaning. Due to its subtle and sophisticated nature, sarcasm is often employed as a tool for trolling or mocking others on social media platforms. Users intentionally exploit the gap between literal meaning and intended sentiment to convey irony, criticism, or humor in a veiled manner, making detection particularly challenging \cite{Tindale1987TheUO, Averbeck2013ComparisonsOI, maynard-greenwood-2014-cares, badlani-etal-2019-ensemble, ghosh-etal-2021-laughing}. With the rise in multimedia content on these platforms, research towards the detection of image-text sarcasm has gained significant attention in recent years.

Understanding and analyzing sarcastic image-text posts poses a significant challenge, as their implicit meaning is not always conveyed directly through the surface-level text or visual content, frequently relying on external contextual information.
For instance, in Figure 1(a), the caption claims a shortage of cabs, while the image clearly depicts an abundance of them. This form of sarcasm is direct and easily recognizable without requiring external knowledge. In Figure 1(b), the caption "time to change the channel for some good news" mocks the idea of escaping a storm warning by simply switching channels. While still indirect, the sarcasm is relatively easy to grasp through description of visual content. However, in Figure 1(c), which suggests Halloween items might appear as early as March or April, relies on knowledge of seasonal retail patterns to detect the sarcasm, which makes it extremely difficult to detect.

In light of the above discussed examples, earlier works \cite{pan-etal-2020-modeling, Liang2021MultiModalSD, tian-etal-2023-dynamic, qin-etal-2023-mmsd2, Tang2024LeveragingGL} rely solely on the image and text inputs without incorporating any external knowledge. While such models can effectively detect sarcasm in cases with explicit incongruities, such as in Figure \ref{fig:intro}(a), they often struggle when deeper contextual understanding is required. To address this, recent studies \cite{cai-etal-2019-multi, Liang2022MultiModalSD, liu-etal-2022-towards-multi-modal} have introduced external knowledge in the form of image captions or object–attribute pairs. Although these cues can help in scenarios like Figure \ref{fig:intro}(b), they are typically shallow and generic, offering surface-level descriptions with limited relevance to the underlying sarcasm. As a result, these approaches fall short in more complex cases that require commonsense reasoning or background knowledge about social or cultural contexts, as illustrated in Figure \ref{fig:intro}(c).

Hence, we argue the need for rich contextual reasoning, where external knowledge is not only verbose but also structured into modality-specific reasoning, for gaining multiple viewpoints from the input. Unlike generic external knowledge used by earlier models, these reasonings should be sarcasm-oriented, offering clear and relevant cues. In designing our approach, we are guided by the following key observations:
\begin{enumerate}
    \item \textbf{Leveraging LVLMs for Rationales:} Large Vision-Language Models (LVLMs), equipped with rich pretrained knowledge, can be prompted to generate explanatory rationales for determining whether an image-text pair is sarcastic. These rationales offer external contextual cues that go beyond the surface-level content of the input.

    \item \textbf{Multimodal dependence of sarcasm:} Sarcasm is not only dependent on the contradiction between the image and text but can also be inferred from individual modalities independently. Therefore, it is crucial to incorporate rationales from the image modality, the text modality, and their combined interpretation.

    \item \textbf{Hallucination and bias in generated rationales:} While LVLM-generated rationales can be informative, they may include hallucinations or factual errors. Relying on them blindly can mislead the model. Therefore, our approach should combine these rationales with cues from the input image-text for more reliable predictions.
\end{enumerate}
To this end, we propose \textbf{MiDRE} (\textbf{Mi}xture of \textbf{D}ual \textbf{R}easoning \textbf{E}xperts), a model designed to explicitly combine external reasoning with internal reasoning from input image-text for multimodal sarcasm detection. To realise observations (1) and (2), we generate rationales from a LVLM using chain-of-thought (CoT) prompting, guiding it to produce step-by-step rationales: first from the image, then the text, and finally from the combined image-text input.

To alleviate the hallucination problem specified in observation (3), MiDRE introduces two complementary expert networks: an internal reasoning expert network (\textit{IRE}) and an external reasoning expert network (\textit{ERE}) within an LLM backbone. The \textit{IRE} considers the raw image-text pair, capturing sarcasm from surface-level incongruities, without relying on external information. To capture sarcastic cues from external knowledge, \textit{ERE} considers the image-text content as well as the rationales generated from the LVLM.  A learnable gating module dynamically fuses the outputs from both experts, enabling MiDRE to adaptively balance between noisy external rationales and cues from the input image-text. This selective integration leads to more robust and interpretable sarcasm detection. MiDRE achieves notable improvements over the best-performing baselines, with accuracy gains of $\Delta$1.84 and $\Delta$1.04 and F1-score gains of $\Delta$1.02 and $\Delta$3.79 on MMSD and MMSD2.0, respectively.
\textbf{Our key contributions are as follows:}
\begin{itemize}
    \item We introduce MiDRE, a novel framework that uses external reasoning and the reasoning based on content, using dual expert networks, with an adaptive and interpretable gating mechanism for identifying sarcasm.
    \item To address the semantic gap of contextual knowledge, we leverage multi-level external rationales (image, text, and combined) generated by a pretrained LVLM with chain-of-thought prompting. 
    \item Through extensive experimentation, we show that MiDRE consistently surpasses existing baselines, achieving higher performance while making it possible to trace and understand the model’s reasoning pathway via gating patterns.
\end{itemize}

\section{Related Work}

\subsection{Image-Text Sarcasm Detection}
Early approaches to sarcasm detection focused solely on text, leveraging lexical features \cite{Joshi2015HarnessingCI,khattri-etal-2015-sentiment, joshi-etal-2016-word,amir-etal-2016-modelling} and sequence modeling techniques \cite{zhang-etal-2016-tweet, poria-etal-2016-deeper, ghosh-etal-2017-role, Agrawal2018AffectiveRF, Agrawal2020LeveragingTO, babanejad-etal-2020-affective, Lou2021AffectiveDG, liu-etal-2022-dual}. The introduction of multimodal sarcasm detection began with handcrafted image-text feature fusion \cite{Schifanella2016DetectingSI}. Subsequent models explored hierarchical fusion \cite{cai-etal-2019-multi}, cross-modal semantic reasoning \cite{xu-etal-2020-reasoning}, and attention mechanisms for capturing intra- and inter-modal incongruities \cite{pan-etal-2020-modeling}. Graph-based methods were employed to model fine-grained token-level relations \cite{Liang2021MultiModalSD, Liang2022MultiModalSD}, while external knowledge incorporation further enhanced performance \cite{liu-etal-2022-towards-multi-modal}. Dynamic routing strategies were later introduced for adaptive inter-modal reasoning \cite{tian-etal-2023-dynamic}. More recent work includes the MMSD2.0 dataset with refined annotations and CLIP-based modeling \cite{qin-etal-2023-mmsd2}, and retrieval-augmented instruction tuning approaches \cite{Tang2024LeveragingGL}.

While prior efforts focus on modeling multimodal incongruity and contrast, often with surface level knowledge, they often lack explicit, sarcasm-oriented reasoning signals. In this work, we leverage structured and verbose reasoning extracted via chain-of-thought (CoT) methods to guide the model to make more robust decisions via dual reasoning expert networks. 

\begin{figure*}[t]
\centering
\includegraphics[width=0.98\textwidth, keepaspectratio]{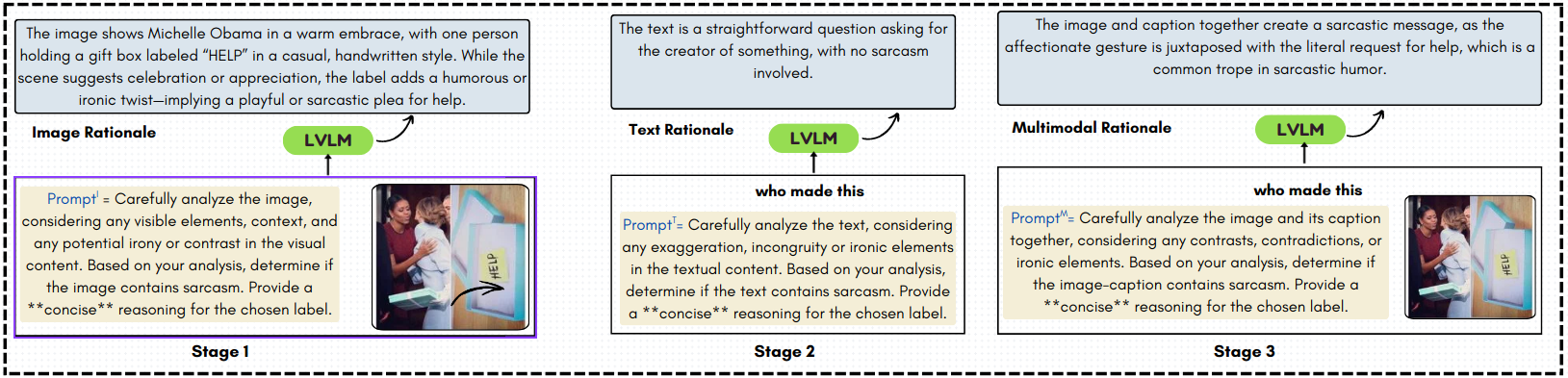} 
\caption{\label{fig:rationale_module}Rationale generation module of MiDRE.}
\end{figure*}

\subsection{Mixture of Experts}
\cite{Jacobs1991AdaptiveMO} proposed the idea of different experts for solving sub-tasks of a larger task. In the domain of multimodal learning, several studies have used MoE approaches for improving multimodal recommendation systems \cite{Lu2015ContentBasedCF, Ma2019SNRSR, Qin2020MultitaskMO}. \cite{Zhang2024CLIPMoETB} used MoE approach to improve the encoding performance of CLIP. \cite{Lin2024MoELLaVAMO} used soft-routers in MoE setting to achieve scaling in LLaVA model. A series of works \cite{Wang2021VLMoUV, Long2023MultiWayAdapaterAL, Shen2023ScalingVM} assigned different experts to handle different modalities. In the domain of multimodal (audio + text + images) analysis of sarcasm, \cite{Yu2023MMoEEM} used different large models as experts in audio, image or text modalities to model their interaction.

In contrast, we propose a novel framework that addresses the problem of sarcasm detection (image+text) from dual perspective, content and external rationales. Our dual expert approach can identify saracsm based on considering the content or external rationales or a mix of both.

\section{Methodology}

\subsection{Task Definition}
The task of image-text sarcasm detection can be framed as a binary classification problem where, given a multimodal input sample pair $x = (T, I)$, where $T$ is the textual content and $I$ is the image, the objective is to assign $x$ a label $y$ from the set $Y = \{yes, no\}$ where $yes$ means sarcastic.

\subsection{Architecture of MiDRE}
\begin{figure*}[t]
\centering
\label{fig:pemcamp}
\includegraphics[width=0.98\textwidth, keepaspectratio]{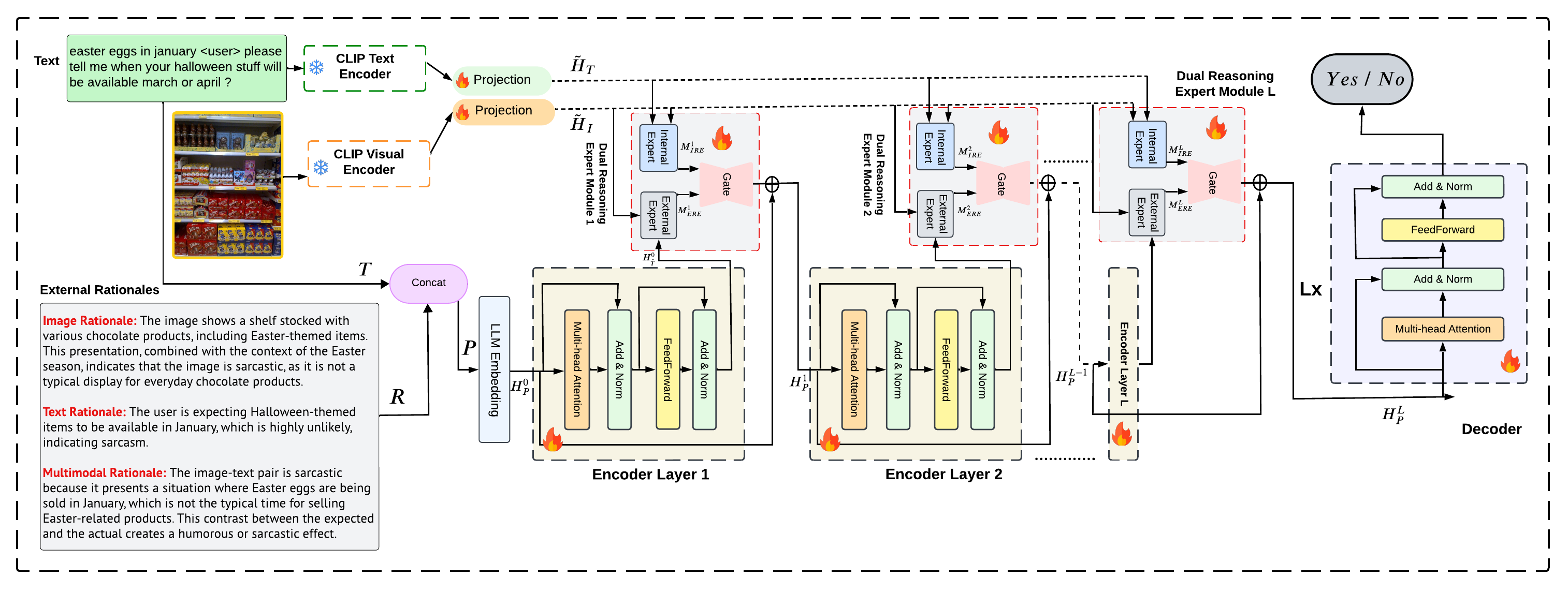} 
\caption{\label{fig:model}Architecture of the proposed MiDRE model.}
\end{figure*}

Our proposed model, MiDRE, detects sarcasm by reasoning from two complementary perspectives: (1) the content itself, and (2) the content enriched with external rationales. As shown in Figure~\ref{fig:model}, the internal reasoning expert \textit{IRE} captures inherent sarcastic cues directly from the input, using visual and textual features extracted via CLIP encoders. To identify sarcasm using both content and external knowledge, we first introduce a Rationale Generation Module that leverages a pre-trained LVLM with CoT prompting to generate structured, modality-specific sarcasm rationales across three views: image, text, and combined image-text. The external reasoning expert \textit{ERE} performs deeper reasoning using LLM-embedded representations of the input text and generated rationales, alongside CLIP-based image embeddings. A learnable gating network then adaptively fuses the outputs of \textit{ERE} and the \textit{IRE}, allowing the model to balance its reliance on external rationales versus direct image-text cues for more accurate and robust sarcasm detection. This combination of \textit{IRE}, \textit{ERE} and gating is incorporated in all encoder layers of an encoder-decoder based LLM to enable dynamic expert selection at multiple levels of representation. This would ensure that both internal and external reasoning cues are effectively leveraged throughout the model’s processing pipeline.

\subsection{Rationale Generation Module}
This module, illustrated in Figure \ref{fig:rationale_module}, is responsible for generating external reasoning or rationales about why a given image-text pair is sarcastic or not, by leveraging the pretrained knowledge of LVLMs. Sarcasm can emerge from either one modality alone or through an interplay between both modalities. To capture these diverse scenarios, we extend the idea of CoT prompting to the multimodal setting by sequentially guiding LVLM to reason in three independent stages: image-only, text-only, and joint image-text. 

We observe that when the LVLM was prompted with both the image and text to generate all three rationales together, it often blended the sources—such as incorporating image details into the text-only rationale, and vice versa. To avoid this cross-modal interference and ensure that each rationale reflects its intended modality, we designed an independent three-stage extraction process. This setup helps maintain clear boundaries between image, text, and combined reasoning. The three-stage process is described below.

\textbf{Stage 1- Image-based rationale generation:} In this stage, the LVLM is prompted to focus exclusively on the image, instructing it to describe the image content and reason whether the visual content indicates sarcasm. This helps capture cues like meme-style content or ironic visual contexts. This stage can be formulated as:
$R^{I} = LVLM(Prompt^{I}, I)$

\textbf{Stage 2- Text-based rationale generation:} Here, the LVLM is prompted to consider only the text to identify signs of sarcasm. This stage isolates the textual sarcasm signal without visual influence. We formulate this stage as:
$R^{T} = LVLM(Prompt^{T}, T)$

\textbf{Stage 3- Multimodal rationale generation:} In the final stage, both image and text are provided as input, and the LVLM is prompted to reason about the relationship between them. The goal of this stage is to identify cross-modal inconsistencies. We can formulate this stage as follows:
$R^{M} = LVLM(Prompt^{M}, I, T)$

We combine the three rationales to form a comprehensive external reasoning input $R = [R^{I}, R^{T}, R^{M}]$. This combined rationale
encapsulates detailed external reasoning signals from each modality and their interaction. \textit{The order of combination is arbitrary, as we observed very little variation in performance during our experiments}.

\subsection{Dual Reasoning Expert Networks}

Our model integrates the \textit{Internal Reasoning Expert} (\textit{IRE}), the \textit{External Reasoning Expert} (\textit{ERE}), and an \textit{adaptive gating} mechanism into \emph{every} encoder layer of the LLM. This layered design enables the model to dynamically combine reasoning signals derived from both original inputs and auxiliary rationales.

\textbf{\textit{External Reasoning Expert (ERE)}:}
Once the rationale $R$ is extracted from the rationale generation module, we construct a rationale-augmented input sample
\[
x = \{I, T, P\}, \quad P = [T, R],
\]
where $I$ is the image, $T$ is the original text, and $P$ is the concatenation of text and rationale. 

The $ERE$ takes as input the image $I$ and the enriched sequence $P$. Image features are extracted using a CLIP visual encoder:
\[
H_{I} = VE(I), \quad H_{I} \in \mathbb{R}^{m \times d_{v}},
\]
where $m$ is the number of image patches. The sequence $P$ contains richer semantic information than $T$ alone, so we pass it through the LLM’s transformer encoder layers to capture deeper contextual dependencies and enable sophisticated integration of external rationale signals. $P$ is embedded by the LLM embedding layer:
\[
H_{P}^{0} = Emb_{\text{LLM}}(P), \quad H_{P}^{0} \in \mathbb{R}^{n \times d_{t}},
\]
where $n$ is the token length of $P$, $d_{t}$ is the LLM embedding dimension, and $H_{P}^{0}$ represents the initial embedding from the LLM that will be fed to the first encoder layer. To align the CLIP visual features with the LLM’s embedding space, we apply a learnable projection:
\[
H_{I^{\prime}} = H_{I} W_{i}.
\]

In encoder layer $i$, the $ERE$ output is computed as:
\[
M_{ERE}^{i} = ERE^{i}(H_{P}^{i-1}, H_{I^{\prime}}),
\]
where $ERE$ is implemented as a cross-attention mechanism:
\[
ERE(H_{P}, H_{I^{\prime}}) = \text{softmax}\!\left(\frac{Q_{P}K_{I^{\prime}}^{T}}{\sqrt{d}}\right) V_{I^{\prime}},
\]
with $\{Q_P, K_{I^{\prime}}, V_{I^{\prime}}\} = \{H_{P}W_{Q},\, H_{I^{\prime}}W_{K},\, H_{I^{\prime}}W_{V}\}$ and $\{W_{Q}, W_{K}, W_{V}\} \in \mathbb{R}^{d \times d_{k}}$.

\textbf{\textit{Internal Reasoning Expert (IRE)}:}
While the $ERE$ leverages rationale-augmented inputs, the $IRE$ is designed to operate \emph{only} on the original image-text pair $(I, T)$, learning semantic associations without external rationale guidance. Since $T$ is a simpler sequence compared to the enriched input $P$, passing it through the LLM encoder is unnecessary and could introduce interference with $P$'s processing. Instead, $T$ is encoded using the CLIP text encoder, which is sufficient to capture its semantic content while keeping the LLM encoder dedicated to the more complex reasoning required for $P$.

We embed $T$ using the CLIP text encoder and project it to the LLM embedding space to get $H_{T^{\prime}}$:
\[
H_{T} = TE(T), \quad H_{T} \in \mathbb{R}^{n \times d_{t}}, \quad H_{T^{\prime}} = H_{T} W_{t}.
\]

The $IRE$ uses the same projected image features $H_{I^{\prime}}$ as the $ERE$, and in layer $i$ computes:
\[
M_{IRE}^{i} = IRE^{i}(H_{T^{\prime}}, H_{I^{\prime}}),
\]
\[
\text{where,} \hspace{0.2cm} IRE(H_{T^{\prime}}, H_{I^{\prime}}) = \text{softmax}\!\left(\frac{Q_{T^{\prime}}K_{I^{\prime}}^{T}}{\sqrt{d}}\right) V_{I^{\prime}},
\]
and $\{Q_{T^{\prime}}, K_{I^{\prime}}, V_{I^{\prime}}\} = \{H_{T^{\prime}}W_{Q},\, H_{I^{\prime}}W_{K},\, H_{I^{\prime}}W_{V}\}$.

\texttt{Design Choice:} $P$ is processed by the LLM encoder and $T$ by the CLIP text encoder, reflecting the higher complexity of $P$ relative to $T$. Reversing this configuration, assigning $T$ to the LLM encoder and $P$ to CLIP text encoder resulted in lower performance, reported in Table \ref{tab:ablation} (row 7) indicated by \texttt{w Swap Enc (P $\leftrightarrow$ T)}.

\textbf{\textit{Adaptive Gating:}}
In many cases, $IRE$ alone suffices for sarcasm detection when the image and text provide strong intrinsic cues. However, certain examples require contextual knowledge captured by $ERE$, while in other cases, $ERE$ may introduce noise. To dynamically balance these contributions, we employ a trainable gating mechanism in each encoder layer $i$.

The gate takes the concatenated outputs of both experts:
\[
G^{i} = \text{softmax}\!\left(W_{g2}^{i} \cdot \text{GELU}\left(W_{g1}^{i}[M_{ERE}^{i};\, M_{IRE}^{i}] \right)\right),
\]
where $W_{g1}^{i} \in \mathbb{R}^{2d_{t} \times d_{b}}$, $W_{g2}^{i} \in \mathbb{R}^{d_{b} \times 2}$, $d_{b} < d$ is the bottleneck dimension, and $G^{i} \in \mathbb{R}^{2}$.

The gated output is then computed as:
\[
M_{G}^{i} = \big(G^{i}[1]\big) \odot M_{ERE}^{i} \;+\; \big(G^{i}[2]\big) \odot M_{IRE}^{i},
\]

We apply a residual connection to obtain the final representation for the next encoder layer:
\[
H_{P}^{i} = H_{P}^{i-1} + M_{G}^{i}.
\]

This design allows the model to selectively attend to internal reasoning, external reasoning, or a context-dependent combination of both at each layer, ensuring robustness to noisy rationales while retaining the benefits of enriched contextual cues.

\subsection{Model Training and Prediction}
The output $H_{P}^{L}$ from the final encoder layer $L$ is fed to the LLM decoder, which generates the label from the label set $\{yes, no\}$. The LLM and the image and text projection layers are trained to minimize the cross-entropy loss between the ground truth label and the generated label. The CLIP visual and text encoders are only used as feature extractors and are frozen during training.
\section{Experiments and Results}

\subsection{Datasets and Experimental Settings}
\begin{figure}[t]
\centering
\includegraphics[width=0.3\textwidth, keepaspectratio]{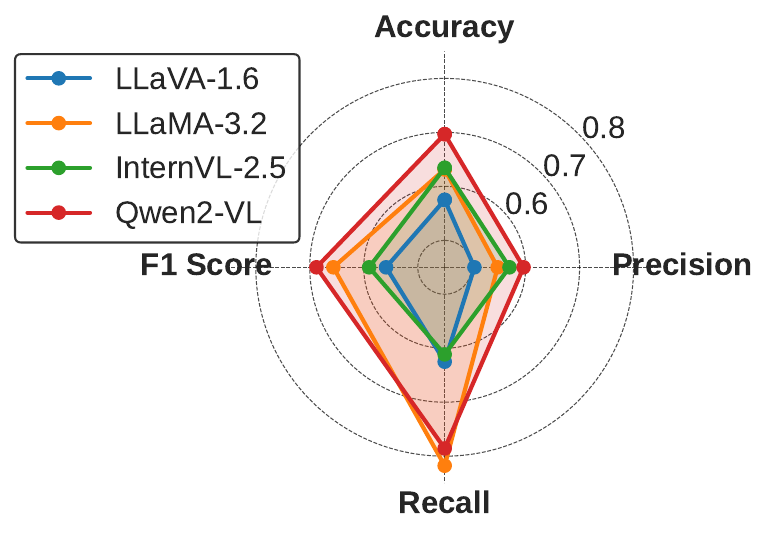} 
\caption{\label{fig:lvlm_comp}Zero-shot Performance of LVLMs on MMSD2.0 test samples.}
\end{figure}

We evaluate our proposed model MiDRE on MMSD2.0 \cite{qin-etal-2023-mmsd2} dataset. We also report the performance on MMSD\cite{cai-etal-2019-multi}, which is a biased dataset in Table \ref{tab:results_mmsd} and discussed in \S \ref{subsec:mmsd_discussion}.
The dataset statistics are reported in Table \ref{tab:stats}. 
We use \emph{flan-t5-large}\footnote{\url{https://huggingface.co/google/flan-t5-large}} as our backbone LLM. 
For generating rationales, we use the LVLM \emph{Qwen2-VL-7B-Instruct}\footnote{\url{https://huggingface.co/Qwen/Qwen2-VL-7B-Instruct}}, due to its better zero-shot performance on our task over other LVLMs as validated in Figure \ref{fig:lvlm_comp}. 
For image and text feature extractors, we use \emph{clip-vit-large-patch14}\footnote{\url{https://huggingface.co/openai/clip-vit-large-patch14}}. We train our model for 20 epochs with Adam optimizer, maintaining a batch size of 12 and a learning rate of 5e-5. We pick the checkpoint with the best validation accuracy for testing. All experiments are carried out on 1xNvidia A100 80GB GPU. 


\begingroup
\renewcommand{\arraystretch}{0.9} 
\setlength{\tabcolsep}{4pt} 
\small
\begin{table}[t]
  \centering
  \caption{Statistics of datasets.}
  \label{tab:stats}
  \begin{tabular}{cccc}
    \toprule
    MMSD/MMSD2.0 & Train & Val & Test \\
    \midrule
    Sentences & 19816/19816 & 2410/2410 & 2409/2409 \\
    Positive & 8642/9572 & 959/1042 & 959/1037 \\
    Negative & 11174/10230 & 1451/1368 & 1450/1372 \\
    \bottomrule
  \end{tabular}
\end{table}

\endgroup

\begingroup
\begin{table}[t]
    \centering
    \small
    \renewcommand{\arraystretch}{0.9}
    \setlength{\tabcolsep}{6pt}
    \caption{Performance comparison of different models on the MMSD2.0 dataset. We reimplement all models. $^{\dagger}$ indicates our method significantly outperforms baselines with $p<0.05$. Best results are in bold.}
    \label{tab:results_mmsd2}
    \begin{adjustbox}{max width=\columnwidth}
    \begin{tabular}{lcccc}
        \toprule
        \textbf{Model} & \textbf{Acc.} & \textbf{P} & \textbf{R} & \textbf{F1} \\
        \midrule
        \multicolumn{5}{c}{\textbf{Text-Only}} \\
        TextCNN \cite{Kim2014ConvolutionalNN} & 71.61 & 64.62 & 75.22 & 69.52 \\
        BiLSTM \cite{Graves20052005SI} & 72.48 & 68.02 & 68.08 & 68.05 \\
        SMSD \cite{Xiong2019SarcasmDW} & 73.56 & 68.45 & 71.55 & 69.97 \\
        RoBERTa \cite{Liu2019RoBERTaAR} & 79.66 & 76.74 & 75.70 & 76.21 \\
        \midrule
        \multicolumn{5}{c}{\textbf{Image-Only}} \\
        ResNet \cite{He2015DeepRL} & 65.50 & 61.17 & 54.39 & 57.58 \\
        ViT \cite{Dosovitskiy2020AnII} & 72.02 & 65.26 & 74.83 & 69.72 \\
        \midrule
        \multicolumn{5}{c}{\textbf{Multimodal}} \\
        HFM \cite{cai-etal-2019-multi} & 70.57 & 64.84 & 69.05 & 66.88 \\
        Att-BERT \cite{pan-etal-2020-modeling} & 80.03 & 76.28 & 77.82 & 77.04 \\
        CMGCN \cite{Liang2022MultiModalSD} & 79.83 & 75.82 & 78.01 & 76.90 \\
        HKE \cite{liu-etal-2022-dual} & 76.50 & 73.48 & 71.07 & 72.25 \\
        DynRT-Net \cite{tian-etal-2023-dynamic} & 71.40 & 71.80 & 72.17 & 71.34 \\
        Multi-view CLIP \cite{qin-etal-2023-mmsd2} & 85.14 & 80.18 & 88.21 & 84.00 \\
        RAG-LLaVA \cite{Tang2024LeveragingGL} & 85.11 & 80.02 & 87.34 & 83.51 \\
        \midrule
        MiDRE (ours) & \textbf{86.18}$^{\dagger}$ & \textbf{86.62}$^{\dagger}$ & \textbf{89.00}$^{\dagger}$ & \textbf{87.79}$^{\dagger}$ \\
        \bottomrule
    \end{tabular}
    \end{adjustbox}
\end{table}
\endgroup

\subsection{Baselines}
We consider three groups of baselines: image-modality, text-modality and multimodal methods. For image-modality, we follow \cite{cai-etal-2019-multi} and use the representations from the pooling layer of  \textbf{ResNet} \cite{He2015DeepRL} and embedding of the [CLS] token from \textbf{ViT} \cite{Dosovitskiy2020AnII}. For text-modality methods, we consider \textbf{TextCNN} \cite{Kim2014ConvolutionalNN}, \textbf{BiLSTM} \cite{Graves20052005SI}, \textbf{SMSD} \cite{Xiong2019SarcasmDW} which uses a low rank bilinear pooling using a self-matching network, and \textbf{RoBERTa}\cite{Liu2019RoBERTaAR}.
For the multimodal methods, we consider approaches designed for multimodal sarcasm detection as baselines, namely, \textbf{HFM} \cite{cai-etal-2019-multi}, \textbf{D\&R Net} \cite{xu-etal-2020-reasoning}, \textbf{Att-BERT} \cite{pan-etal-2020-modeling}, \textbf{InCrossMGs} \cite{Liang2021MultiModalSD}, \textbf{CMGCN} \cite{Liang2022MultiModalSD}, \textbf{HKE} \cite{liu-etal-2022-towards-multi-modal}, \textbf{DynRT-Net} \cite{tian-etal-2023-dynamic}, \textbf{Multi-view CLIP} \cite{qin-etal-2023-mmsd2} and \textbf{RAG-LLaVA} \cite{Tang2024LeveragingGL}.
We do not show the results of \textbf{InCrossMGs} and \textbf{D\&R Net} on MMSD2.0 due to the unavailability of their codes. 

\begingroup
\begin{table}[t]
    \centering
    \small
    \renewcommand{\arraystretch}{0.9}
    \setlength{\tabcolsep}{6pt}
    \caption{Performance comparison of different models on the MMSD dataset. We reimplement all models. $^{\dagger}$ indicates our method significantly outperforms baselines with $p<0.05$. Best results are highlighted in bold.}
    \label{tab:results_mmsd}
    \begin{adjustbox}{max width=\columnwidth}
    \begin{tabular}{lcccc}
        \toprule
        \textbf{Model} & \textbf{Acc.} & \textbf{P} & \textbf{R} & \textbf{F1} \\
        \midrule
        \multicolumn{5}{c}{\textbf{Text-Only}} \\
        TextCNN \cite{Kim2014ConvolutionalNN} & 80.03 & 74.29 & 76.39 & 75.32 \\
        BiLSTM \cite{Graves20052005SI} & 81.90 & 76.66 & 78.42 & 77.53 \\
        SMSD \cite{Xiong2019SarcasmDW} & 80.90 & 76.46 & 75.18 & 75.82 \\
        RoBERTa \cite{Liu2019RoBERTaAR} & 93.97 & 90.39 & 94.59 & 92.45 \\
        \midrule
        \multicolumn{5}{c}{\textbf{Image-Only}} \\
        ResNet \cite{He2015DeepRL} & 64.76 & 54.41 & 70.80 & 61.53 \\
        ViT \cite{Dosovitskiy2020AnII} & 67.83 & 57.93 & 70.07 & 63.40 \\
        \midrule
        \multicolumn{5}{c}{\textbf{Multimodal}} \\
        HFM \cite{cai-etal-2019-multi} & 83.44 & 76.57 & 84.15 & 80.18 \\
        D\&R Net \cite{xu-etal-2020-reasoning} & 84.02 & 77.97 & 83.42 & 80.60 \\
        Att-BERT \cite{pan-etal-2020-modeling} & 86.05 & 80.87 & 85.08 & 82.92 \\
        InCrossMGs \cite{Liang2021MultiModalSD} & 86.10 & 81.38 & 84.36 & 82.84 \\
        CMGCN \cite{Liang2022MultiModalSD} & 86.54 & - & - & 82.73 \\
        HKE \cite{liu-etal-2022-dual} & 87.36 & 81.84 & 86.48 & 84.09 \\
        DynRT-Net \cite{tian-etal-2023-dynamic} & 93.59 & 93.06 & 93.60 & 93.31 \\
        Multi-view CLIP \cite{qin-etal-2023-mmsd2} & 88.33 & 82.66 & 88.65 & 88.55 \\
        RAG-LLaVA \cite{qin-etal-2023-mmsd2} & 89.97 & 89.26 & 89.58 & 89.42 \\
        \midrule
        MiDRE (ours) & \textbf{95.81}$^{\dagger}$  & \textbf{93.11}$^{\dagger}$ & \textbf{93.84}$^{\dagger}$ & \textbf{93.47}$^{\dagger}$ \\
        \bottomrule
    \end{tabular}
    \end{adjustbox}
\end{table}
\endgroup

\subsection{Main Results}
Following \cite{cai-etal-2019-multi}, we use Accuracy (Acc.), macro-average precision (P), macro-average recall (R), and macro-average F1 (F1) to report the performance of our model in Table \ref{tab:results_mmsd2}. Our main observations are follows: (1) \textbf{MiDRE outperforms all SOTA baselines in MMSD2.0}: On MMSD2.0 dataset, MiDRE outperforms all baseline across unimodal and multimodal settings. Specifically it ourpeforms the best SOTA multimodal baseline Multi-view CLIP by good margins across metrics (Acc: $\uparrow 1.04\%$, P: $\uparrow 6.44\%$, R: $\uparrow 0.79\%$, F1: $\uparrow 3.79\%$ ). This proves the efficacy of using structured rationales and having dual reasoning paths to identify sarcasm. 

(2) \textbf{MiDRE outperforms all multimodal baselines that use external knowledge:}
We notice that our method significantly outperforms baselines like HFM \cite{cai-etal-2019-multi} (Acc: $\uparrow 15.61\%$ in MMSD2.0) which uses image attributes generated using ResNet, CMGCN \cite{Liang2022MultiModalSD} (Acc: $\uparrow 6.35\%$ in MMSD2.0) which utilizes object-attribute pairs from a detectron module and HKE \cite{liu-etal-2022-towards-multi-modal} (Acc: $\uparrow 9.68\%$ in MMSD2.0) which incorporates image caption, image attributes and adjective-noun pairs from image. This highlights the importance of leveraging structured, verbose and sarcasm-oriented knowledge.

\subsection{Results on Biased MMSD Dataset}
\label{subsec:mmsd_discussion}
MMSD is a text-biased dataset, as shown by \citet{qin-etal-2023-mmsd2}, due to strong textual cues like hashtags and emojis that predominantly occur in one class. Text-only models can exploit these cues without considering visual input, often outperforming multimodal models. For completeness, we evaluate MiDRE on MMSD. As shown in Table \ref{tab:results_mmsd}, the higher performance of RoBERTa over multimodal baselines highlights this bias. Nonetheless, MiDRE consistently outperforms all unimodal and multimodal models by significant margins, demonstrating robust reasoning ability.

\begingroup
\begin{table}[t]
\centering
\small
\caption{Performance comparison of MiDRE with LVLMs on MMSD2.0. Best scores are in \textbf{bold}.}
\label{tab:midre_lvlm}
\setlength{\tabcolsep}{4pt}
\begin{adjustbox}{width=0.8\columnwidth}
\scriptsize
\begin{tabular}{lcccc}
\toprule
\textbf{Model} & \textbf{Acc} & \textbf{P} & \textbf{R} & \textbf{F1} \\
\midrule
MiDRE (Ours)        & \textbf{86.18} & \textbf{86.62} & 89.00 & \textbf{87.79} \\
Qwen-2-VL-7B \cite{Yang2024Qwen2TR}           & 85.71 & 83.36 & 86.20 & 84.69 \\
LLaMA-3.2-11B \cite{Dubey2024TheL3}       & 84.64 & 80.20 & 85.26 & 82.56 \\
LLaVA-1.6-7B \cite{Liu2023ImprovedBW}        & 85.38 & 77.48 & \textbf{93.15} & 84.54 \\
InternVL-2.5-8B \cite{Chen2024ExpandingPB}       & 85.26 & 78.21 & 91.13 & 84.12 \\
\bottomrule
\end{tabular}
\end{adjustbox}
\end{table}
\endgroup

\begingroup

\begin{table}[t]
    \centering
    \renewcommand{\arraystretch}{1.15}

    \caption{Ablation of our model MiDRE. w/o means without that specific component.} 
    \label{tab:ablation}
    \begin{adjustbox}{width=0.98\columnwidth}
    \begin{tabular}{lcccc|cccc}
        \toprule
        \multirow{2}{*}{\textbf{MiDRE}} & \multicolumn{4}{c|}{\textbf{MMSD}} & \multicolumn{4}{c}{\textbf{MMSD2.0}} \\
        & \textbf{Acc.} & \textbf{P} & \textbf{R} & \textbf{F1} & \textbf{Acc.} & \textbf{P} & \textbf{R} & \textbf{F1} \\
        \midrule
         Only Mul. Rationale & 90.13 & 87.35 & 90.41 & 88.85 & 84.06 & 80.51 & 86.69 & 83.48\\
        Only Txt+Img Rationale & 90.67 & 89.21  & 88.37 & 88.78 &  84.68& 82.11 & 82.35 & 82.23\\
          w/o ERE & 89.09 & 85.82 & 89.11 & 87.43 & 82.27 & 74.51 & 89.39 & 81.28\\
          w/o IRE & 91.06 & 89.62 & 88.45 & 89.03 & 84.06 & 80.03 & 83.89 & 81.92\\
          w/o Gate & 90.55 & 85.13 & 89.97 & 87.48 & 84.84 & 78.96 & 88.33 & 83.38\\
          w Linear Gate & 91.11 & 85.82 & 90.02 & 87.86 & 84.89 & 79.17 & 88.52 & 83.58\\
        w Swap Enc (P $\leftrightarrow$ T) & 91.62 & 89.23 & 88.75 & 88.99 & 82.13 & 81.62 & 80.20 & 80.90 \\
          
          \midrule
        MiDRE & 95.81  & 93.11 & 93.84 & 93.47 & 86.18 & 86.62 & 89.00 & 87.79 \\
        \bottomrule
      
    \end{tabular}
    \end{adjustbox}
\end{table}
\endgroup

\begin{figure}[t]
\centering
\includegraphics[width=0.32\textwidth, height=0.25\textwidth]{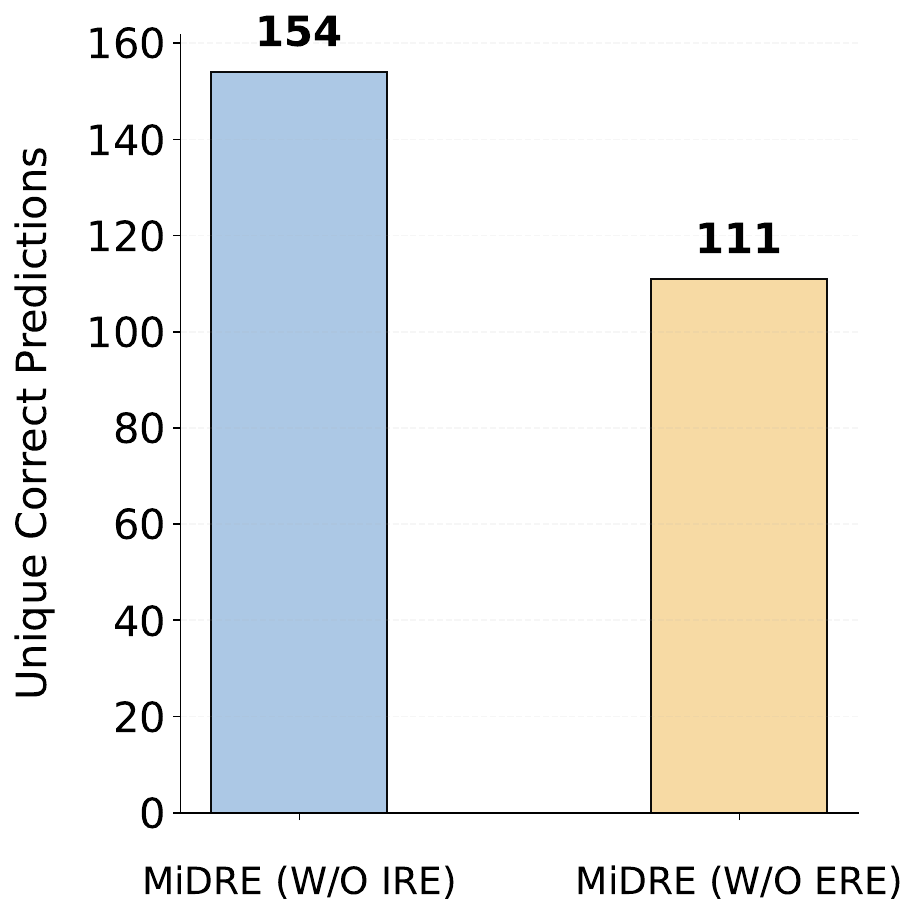} 
\caption{\label{fig:venn}Unique test samples correctly classified by MiDRE(w/o IRE) vs MiDRE(w/o ERE).}
\end{figure}

\section{Comparison with LVLMs}
We compare the performance of MiDRE with four SOTA large vision-language models (LVLMs) by LoRA finetuning them with rank 8, due to resource constraints. The results in Table \ref{tab:midre_lvlm} show that MiDRE consistently outperforms all LVLMs across most metrics. While LLaVA-1.6-7B records the highest recall (93.15\%), its lower precision leads to a weaker F1 than MiDRE. This suggests that MiDRE not only identifies relevant instances effectively but also minimizes false positives, making it more robust. At the same time, it reveals how the capabilities of such smaller models can be significantly amplified by strategically leveraging contextual understanding from large vision-language models.

\begin{figure*}[t]
\centering
\includegraphics[width=1.0\textwidth, keepaspectratio]{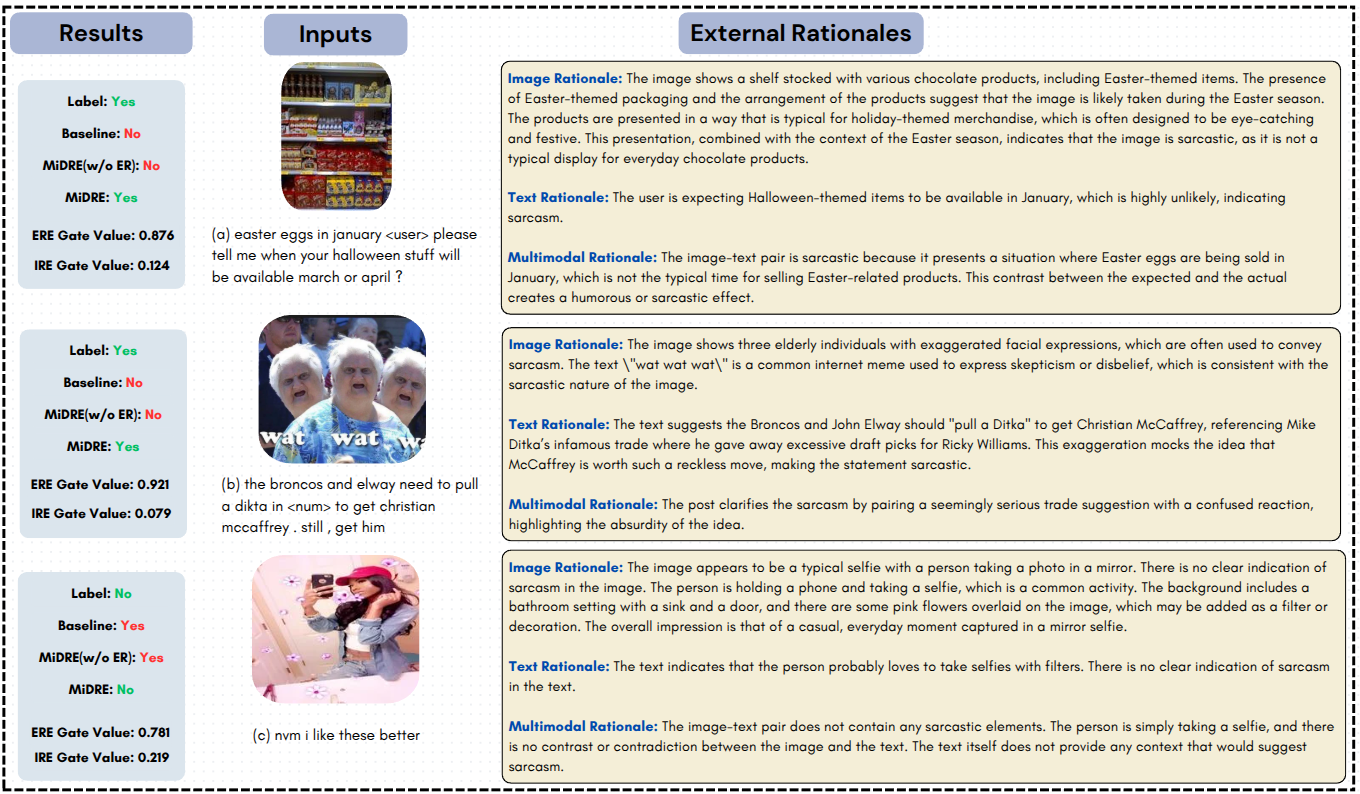} 
\caption{\label{fig:analysis1}Examples demonstrating the importance of external rationales for sarcasm identification. 'Baseline' indicates Multi-view CLIP (the best performing baseline for MMSD2.0), ERE(IRE) Gate Values represent the importance given to external (internal) expert module by the gating network in MiDRE.}
\end{figure*}

\section{Ablation}
To further prove the efficacy of our approach, we answer the following research questions:
(1) \textit{What is the impact of having only one expert network?} 
(2) \textit{What advantage does the gating network provide? } 
(3) \textit{Do we need rationales from all of image, text and multimodal views?} 

\subsection{Single Expert Network}
To better understand the contributions of \textit{IRE} and \textit{ERE}, we perform an ablation study on MiDRE by selectively removing each component (Table \ref{tab:ablation}, rows 3 \& 4). Removing the \textit{ERE} module (\textit{w/o ERE}) leads to a significant drop in accuracy (↓6.72\% on MMSD, ↓3.91\% on MMSD2.0), highlighting the importance of external contextual rationales that are not directly inferable from the input. Conversely, removing \textit{IRE} (\textit{w/o IRE}) also reduces performance (↓4.75\% on MMSD, ↓2.12\% on MMSD2.0), showing that \textit{ERE} alone struggles to compensate for noisy rationales. 

Interestingly, from Figure \ref{fig:venn}, we can observe that \textit{w/o ERE} (only considering \textit{IRE}) correctly classifies 111 unique samples and \textit{w/o IRE} (only considering \textit{ERE}) classifies 154 different ones, suggesting that both experts offer complementary and independently valuable capabilities. Moreover, the larger performance drop without \textit{ERE} indicates its dominant role, but \textit{IRE} remains essential for grounding predictions in the raw image-text content.

\begin{figure}[t]
\centering
\adjustbox{max width=0.7\columnwidth}{
\begin{tabular}{@{}p{0.48\linewidth}p{0.48\linewidth}@{}}
\includegraphics[width=\linewidth, height=0.5\linewidth]{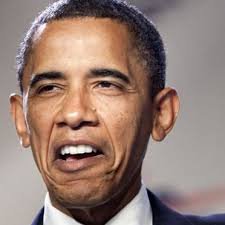} &
\includegraphics[width=\linewidth, height=0.5\linewidth]{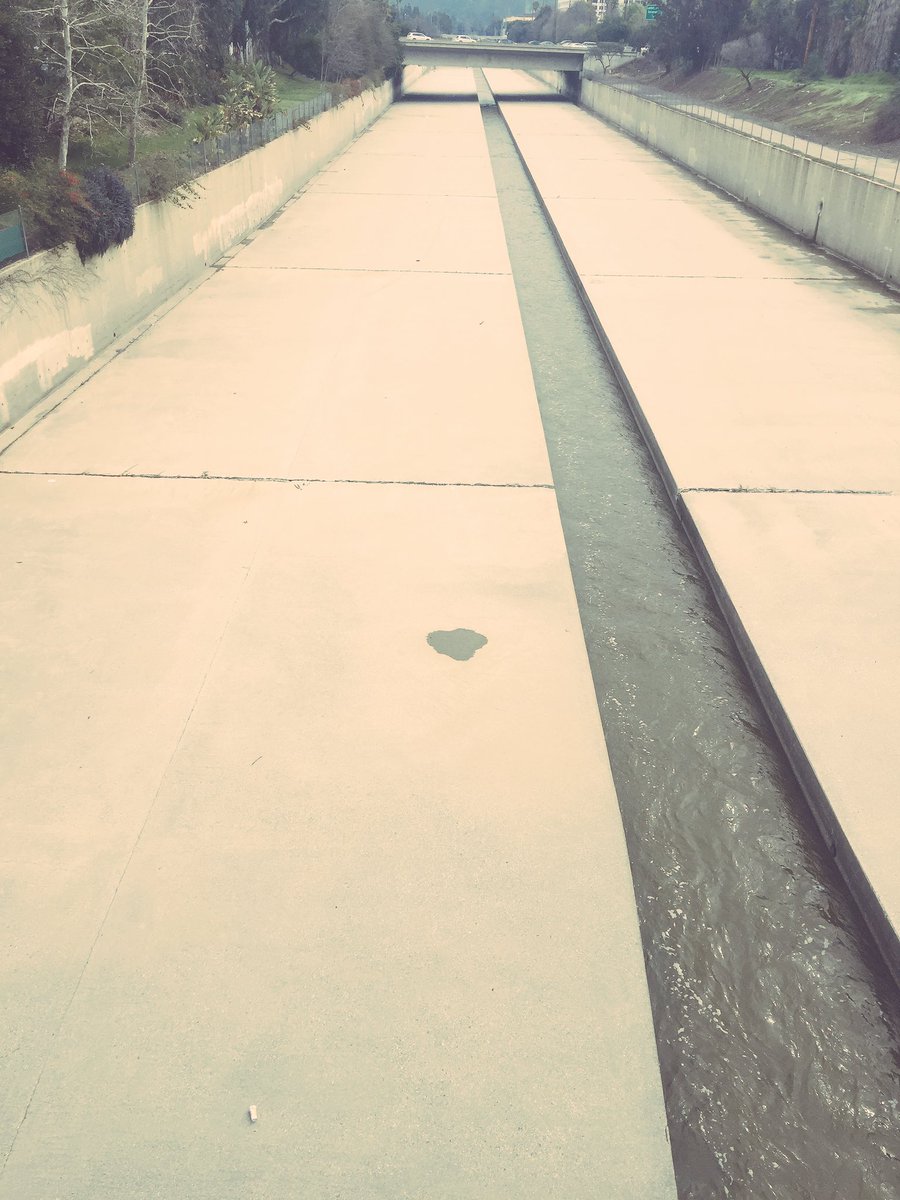} \\
\large\textbf{Text:} \# bringbackobama because who needs the constitution anyway &
\large\textbf{Text:} the beautiful los angeles river-why isn't this a wonder of the world already? \\
\large\textbf{Label:} Sarcastic &
\large\textbf{Label:} Sarcastic \\
\large\textbf{Type:} External &
\large\textbf{Type:} Internal \\
[8pt]
\includegraphics[width=\linewidth, height=0.6\linewidth]{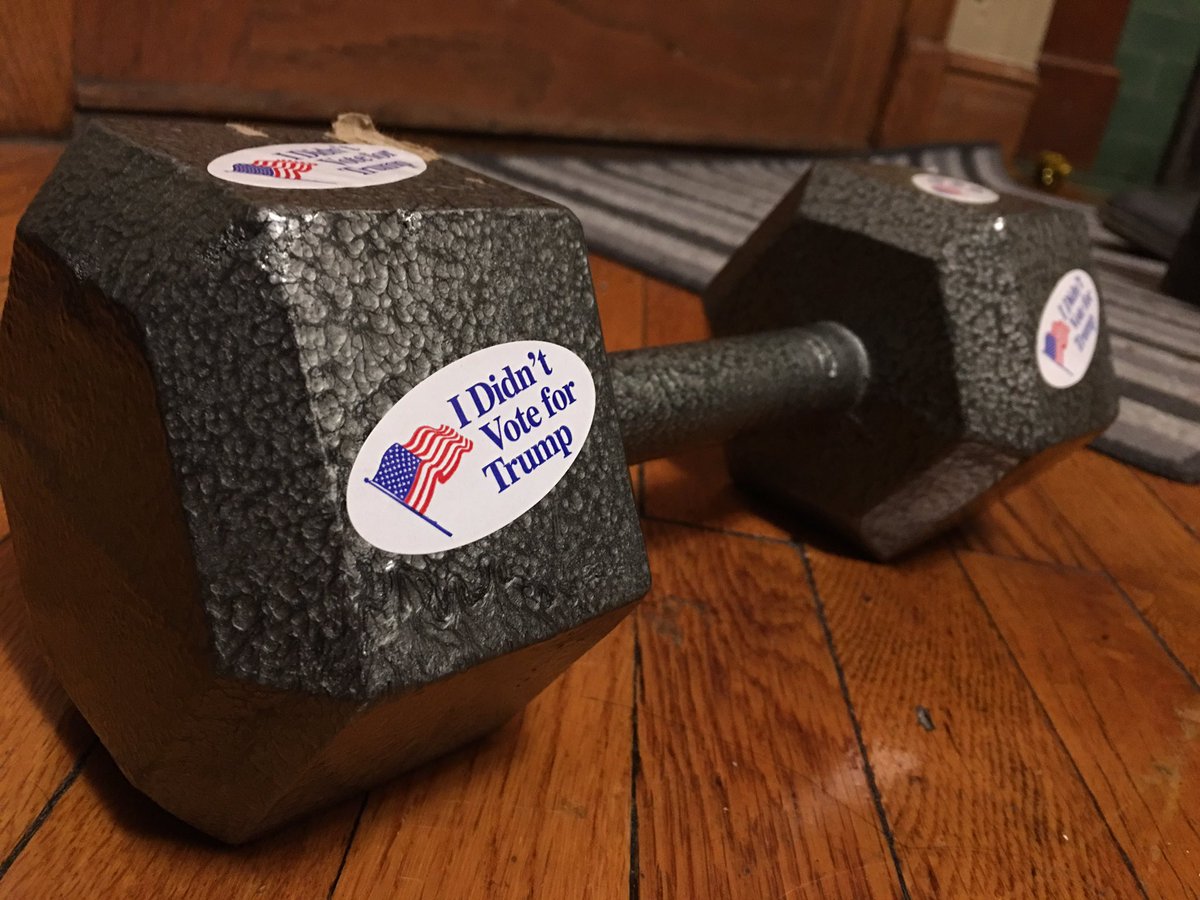} &
\includegraphics[width=\linewidth, height=0.5\linewidth]{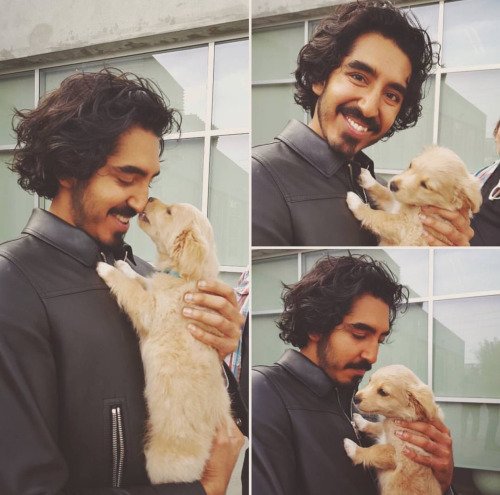} \\
\large\textbf{Text:} radicalizing my biceps! &
\large\textbf{Text:}  i 'm crying this is so cute \\
\large\textbf{Label:} Non-sarcastic &
\large\textbf{Label:} Non-sarcastic \\
\large\textbf{Type:} External &
\large\textbf{Type:} Internal
\end{tabular}
}
\caption{
Representative examples. 
Top row: \textit{internal} vs. \textit{external-knowledge} sarcastic cases. 
Bottom row: \textit{internal} vs. \textit{external-knowledge} non-sarcastic cases.
}
\label{fig:gate_examples}
\end{figure}

\begin{figure}[t]
\centering
\includegraphics[width=0.38\textwidth, keepaspectratio]{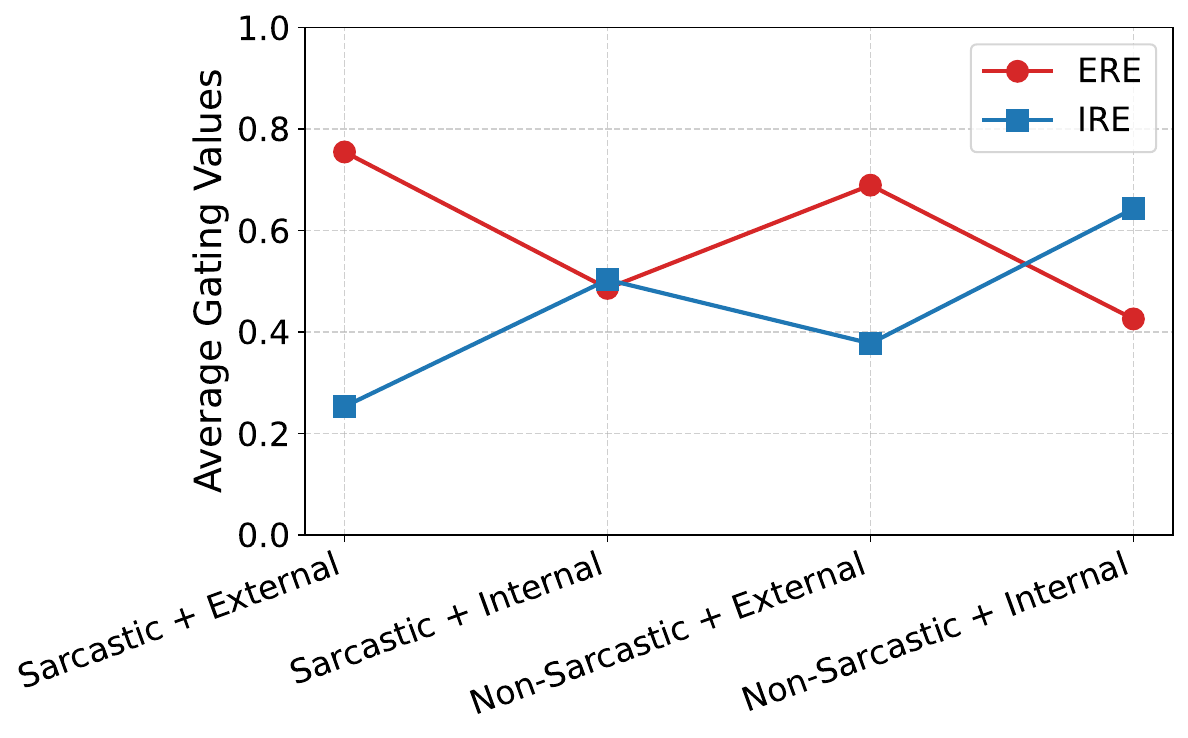} 
\caption{\label{fig:gating_distribution}Gating values distribution across 4 categories of samples. External means samples that heavily rely on external knowledge, while internal represent those classes of samples that do not require external context.}
\end{figure}

\begin{figure*}[t]
\centering
\includegraphics[width=1.0\textwidth, keepaspectratio]{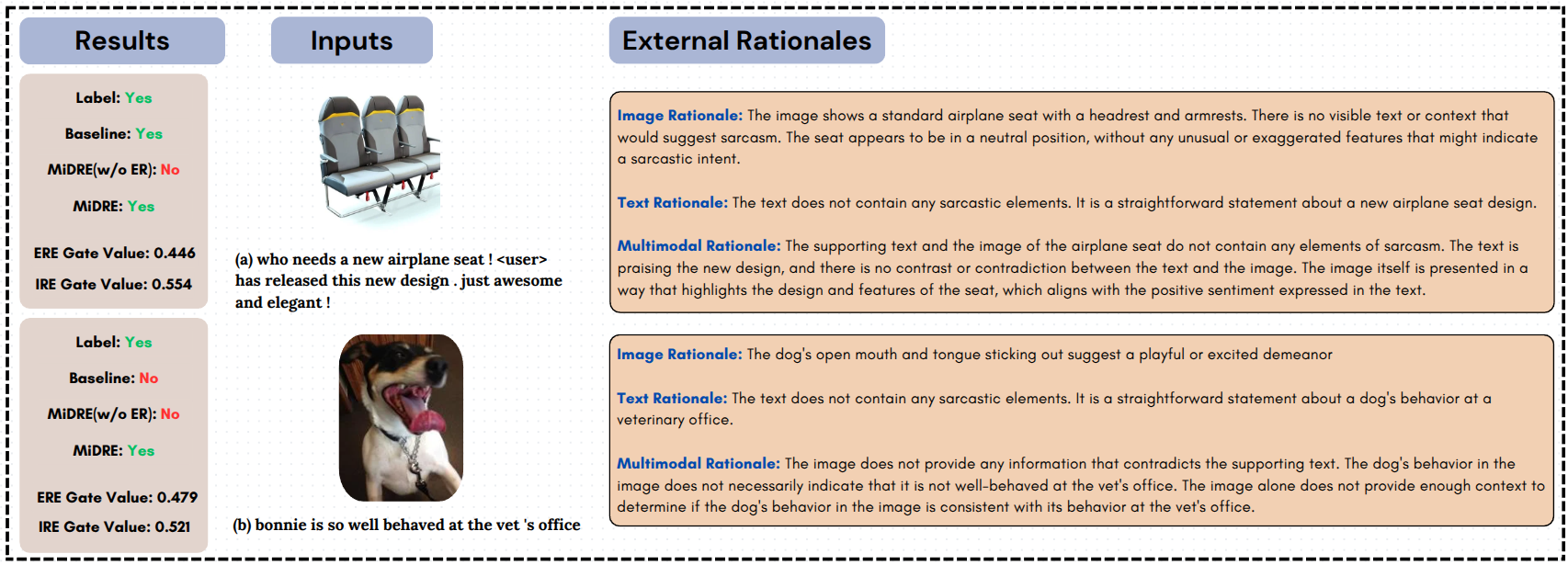} 
\caption{\label{fig:analysis2}Examples demonstrating how noisy rationales can provide contextual cues. 'Baseline' indicates Multi-view CLIP (the best performing baseline for MMSD2.0), ERE(IRE) Gate Values represent the importance given to external (internal) expert module by the gating network in MiDRE.}
\end{figure*}

\subsection{Impact of Gating Network}
To understand the impact of the gating network, we ablate by removing it from MiDRE,  reported in Table \ref{tab:ablation} as \textit{w/o Gate}. In this case, we simply add the output from \textit{IRE} and \textit{ERE}. We notice that performance drops notably (Acc: $\downarrow 5.26\%$ in MMSD and $\downarrow 1.34\%$ in MMSD2.0). This decrease in performance indicates that the gating network plays a crucial role in controlling the interaction between the \textit{IRE} and \textit{ERE}. The adaptive gating ensures that the model relies on the most informative reasoning path to make a decision. In cases where it can't rely on one single reasoning path, it happens to look at both reasoning paths to arrive at a decision, which is described in detail in Section \ref{sec:analysis}. We also validate our design choice of using a bottleneck architecture for the gating mechanism by comparing it against a simpler linear gate, as reported in Table \ref{tab:ablation} (\textit{w Linear Gate}). The bottleneck gate consistently performs better, suggesting that it facilitates more expressive interactions between the expert outputs.

\subsection{Need of All Three Rationales}
To validate the necessity of all three rationales (text, image and multimodal) in MiDRE, we conduct a systematic ablation study presented in Table \ref{tab:ablation} (rows 1 \& 2). When only the multimodal rationale is used, the model's performance drops significantly,(Acc: $\downarrow 5.68\%$ in MMSD and $\downarrow 2.12\%$ in MMSD2.0), indicating that relying solely on the multimodal rationale is insufficient. Similarly, when only the text and image rationale is used, the performance drops (Acc: $\downarrow 5.14\%$ in MMSD and $\downarrow 1.5\%$ in MMSD2.0), compared to the full model. In contrast, the full MiDRE model, which combines the image, text and multimodal rationales achieves the highest performance, emphasizing the necessity of knowledge from all three views.

\section{Analysis}\label{sec:analysis}

\subsection{Gating Values Distribution}
\label{subsec:gating_values_distribution}
To assess whether the gating network prioritizes the \textit{ERE} or \textit{IRE} according to genuine reasoning needs rather than spurious cues, we perform a controlled experiment. We curated 100 test samples (50 sarcastic, 50 non-sarcastic). Within each class, we further identified 25 examples that require external knowledge (e.g., commonsense facts, cultural references, or world events) and 25 examples that can be resolved purely through internal image–text incongruity. The representative examples of each group is shown in Fig. \ref{fig:gate_examples}. For each group, we recorded the gate's softmax weights. We observe in Fig \ref{fig:gating_distribution} that samples needing external knowledge showed consistently higher \textit{ERE} weights across sarcasm labels, while internal-only cases favored \textit{IRE}, suggesting that the gate aligns with reasoning needs rather than superficial cues.

\subsection{Perturbation Analysis of Rationales}
To assess whether MiDRE leverages the external rationales meaningfully rather than ignoring them or learning spurious correlations, we perform a controlled perturbation experiment. We select the same 50 samples (25 sarcastic and 25 non-sarcastic) used in \S \ref{subsec:gating_values_distribution}, which require external knowledge. For each sample, we replace its original rationale with a rationale randomly drawn from another sample in this set (ensuring no self-substitution), while keeping the image and text unchanged. We then re-run the full pipeline and measure: (1) classification accuracy before and after perturbation, and (2) the gate's value for ERE, before and after perturbation.

Table~\ref{tab:pert_full} shows the effect of perturbing external rationales on classification accuracy and the average \textit{ERE} gating weight. Across all external-knowledge samples, accuracy drops by $\downarrow 4.7\%$, with a slightly larger decrease for sarcastic cases ($\downarrow 5.2\%$) than for non-sarcastic cases ($\downarrow 4.2\%$), reflecting greater reliance on external knowledge when sarcasm is present. The mean gate value for \textit{ERE} also decreases across groups (e.g., $\downarrow 0.06$ for sarcastic samples), indicating that the gating network appropriately down-weights the \textit{ERE} when its reasoning is unreliable. However, since these posts genuinely require external knowledge, this reduction in reliance leads to performance degradation, demonstrating that the gate’s behavior aligns with reasoning needs rather than spurious cues.

\begingroup
\begin{table}[t]
\centering
\caption{Perturbation analysis of the external rationales. Negative $\Delta$ shown in \textcolor{red!70!black}{red}.}
\label{tab:pert_full}
\setlength{\tabcolsep}{4pt}
\resizebox{\columnwidth}{!}{%
\scriptsize
\begin{tabular}{lcccccc}
\toprule
\multirow{2}{*}{Group} & \multirow{2}{*}{\#} & \multicolumn{3}{c}{Accuracy (\%)} & \multicolumn{2}{c}{\textit{ERE Gate Value}} \\
\cmidrule(lr){3-5} \cmidrule(lr){6-7}
 & & Orig & Pert & $\Delta$ & Orig & Pert \\
\midrule
All ext.-knowledge & 50 & 88.3 & 83.6 & \textcolor{red!70!black}{$-4.7$} & 0.70 & 0.65 \\
Sarcastic & 25 & 89.3 & 84.1 & \textcolor{red!70!black}{$-5.2$} & 0.74 & 0.68 \\
Non-sarcastic & 25 & 87.4 & 83.2 & \textcolor{red!70!black}{$-4.2$}  & 0.67 & 0.62 \\
\bottomrule
\end{tabular}%
}
\end{table}
\endgroup

\subsection{Need of External Rationale}
External reasoning is crucial in sarcasm detection, particularly when textual and visual cues alone fail to establish contradiction. For instance, in Figure \ref{fig:analysis1}(a), the image shows Easter-themed products on a shelf, while the text questions when Halloween products will arrive, implying that Easter items are being sold too early. A model without external knowledge may struggle to detect sarcasm since the image-text itself does not explicitly indicate any contradiction. However, external reasoning clarifies that Easter eggs appearing in January is unusual, reinforcing the sarcasm. MiDRE successfully classifies this instance as sarcastic, with a high gate value for \textit{ERE} module (0.876), while the baseline and MiDRE without \textit{ERE} fail. In Figure \ref{fig:analysis1}(b), the text references a famous sports trade (Mike Ditka’s excessive trade for Ricky Williams) to suggest that the Broncos should do something similarly extreme. The accompanying image with confused expressions enhances the sarcasm. Without knowledge of the Ditka trade, the sarcasm might be missed, as the text could be interpreted as a genuine suggestion. External reasoning bridges this gap by explaining the historical context, leading MiDRE to detect sarcasm correctly (\textit{ERE} gate value: 0.921). In the case of Figure \ref{fig:analysis1}(c), both the image and text lack strong sarcastic elements. The image depicts a casual mirror selfie, and the text states that the person probably takes selfies with filters. MiDRE assigns a high \textit{ERE} gate value (0.781), suggesting that the \textit{IRE} module might be uncertain due to the overtly positive tone of the text. However, \textit{ERE} helps by recognizing that there is no cultural or contextual contradiction that would indicate sarcasm.

\subsection{Impact of Noisy Rationale}
External rationales, even when noisy or indirect, enhance the \textit{IRE} module by providing valuable contextual cues for sarcasm detection. For example, in Figure \ref{fig:analysis2}(a), MiDRE without \textit{ERE} fails to detect sarcasm, while MiDRE succeeds by leveraging the image rationale, which highlights the neutral design of the seat. This neutrality amplifies the exaggerated praise in the text, making the sarcasm more apparent. Similarly, in \ref{fig:analysis2}(b), although the image rationale doesn’t directly contradict the caption, it describes the dog’s playful behavior, subtly clashing with the claim of it being "well-behaved" at the vet. In both cases, rationales offers factual descriptions that help \textit{IRE} identify subtle incongruities, demonstrating the complementary role of external rationales in improving sarcastic intent recognition.

\begin{figure}[t]
\centering
\includegraphics[width=0.47\textwidth,height=0.63\textwidth]{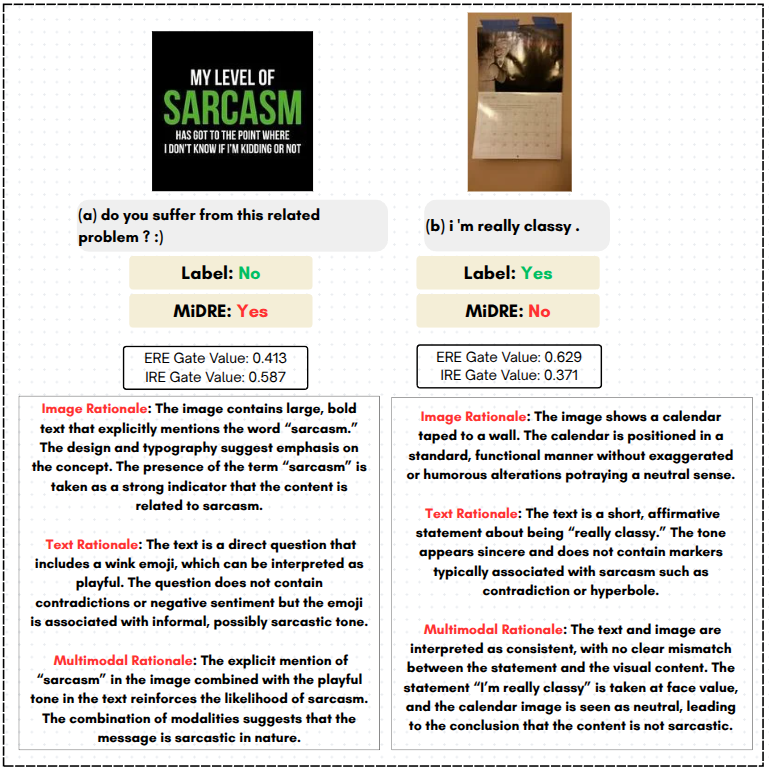} 
\caption{\label{fig:error}Examples of misclassified samples}
\end{figure}

\section{Error Analysis}

Our model, MiDRE, fails in scenarios where both the \textit{ERE} and \textit{IRE} produce weak or misleading signals. In Figure \ref{fig:error}(a), the ERE gate value (0.413) is lower than the IRE gate value (0.587), indicating heavier reliance on internal reasoning. However, the internal reasoning is biased by the explicit presence of the word “sarcasm” in the image text, leading to an incorrect prediction. In Figure \ref{fig:error}(b), the ERE gate value (0.629) outweighs the IRE gate value (0.371), suggesting the model leaned more on external reasoning. Here, the humor comes from the visual-textual contradiction between the caption “I’m really classy” and the taped wall calendar. The internal reasoning misses this subtle contradiction, while external reasoning fails to connect the visual setup with its implied sarcasm. Consequently, the model predicts this post as not sarcasm.

\section{Conclusion}
This work addresses the limitations of existing approaches that use shallow external knowledge, which is often surface-level and generic in nature. We propose MiDRE, a novel framework that combines internal reasoning over image-text pairs with external reasoning guided by structured rationales generated through CoT prompting to a LVLM. MiDRE leverages a learnable gating mechanism to dynamically integrate both reasoning pathways, allowing the model to adaptively rely on commonsense and contextual knowledge when necessary. Extensive experiments on multiple benchmarks demonstrate that MiDRE achieves state-of-the-art performance and provides more interpretable and robust sarcasm detection, especially in complex scenarios that require understanding beyond literal visual or textual cues. This highlights the importance of structured external rationales and expert-guided reasoning in advancing multimodal understanding tasks like sarcasm detection.

\section*{Ethical Considerations}
This work on multimodal sarcasm detection is conducted entirely on publicly available datasets, with all private, personally identifiable content removed during preprocessing to ensure compliance with data protection and ethical research guidelines. The developed model is intended solely for academic and research purposes, and is not optimized, validated, or certified for use in production systems.
While the model can provide insights into sarcasm patterns in multimodal data, its outputs should be interpreted with caution, as sarcasm detection is inherently subjective and may be influenced by cultural, contextual, and individual differences. We strongly discourage the use of this model in high-stakes or sensitive decision-making contexts, including but not limited to legal, political, or law enforcement scenarios.


\appendix

\bibliographystyle{ACM-Reference-Format}
\bibliography{sample-base}

\appendix


\end{document}